# Alpha Excel Benchmark

David A. Noever
PeopleTec, Inc.
Huntsville, AL
david.noever@peopletec.com

Forrest McKee
PeopleTec, Inc.
Huntsville, AL
forrest.mckee@peopletec.com

**ABSTRACT**

This study presents a novel benchmark for evaluating Large Language Models (LLMs) using challenges derived from the Financial Modeling World Cup (FMWC) Excel competitions. We introduce a methodology for converting 113 existing FMWC challenges into programmatically evaluable JSON formats and use this dataset to compare the performance of several leading LLMs. Our findings demonstrate significant variations in performance across different challenge categories, with models showing specific strengths in pattern recognition tasks but struggling with complex numerical reasoning. The benchmark provides a standardized framework for assessing LLM capabilities in realistic business-oriented tasks rather than abstract academic problems. This research contributes to the growing field of AI benchmarking by establishing proficiency among the 1.5 billion people who daily use Microsoft Excel as a meaningful evaluation metric that bridges the gap between academic AI benchmarks and practical business applications.

**INTRODUCTION**
The recent rapid advancement of Large Language Models (LLMs) has sparked interest in developing specialized benchmarks to evaluate their capabilities across various domains. While existing benchmarks often focus on natural language understanding, programming, or reasoning abilities in abstract contexts, there remains a notable gap in benchmarks that assess performance on practical business tasks (Brown et al., 2020). Microsoft Excel, being one of the most widely used business software tools globally, presents an opportunity to create tasks that simultaneously test multiple dimensions of LLM capabilities, including numerical reasoning, pattern recognition, rule comprehension, file conversion, and problem-solving strategies.

The Financial Modeling World Cup (FMWC), established in 2020, has emerged as a premier global competition testing advanced Excel skills through creative challenges that range from financial modeling to game simulations implemented in spreadsheets (Grigolyunovich, 2022). The competition has evolved into a significant esports phenomenon with substantial prize pools and international participation, establishing a rich repository of expert-designed Excel challenges with varying difficulty levels and skill requirements (FMWC, 2024; Sheedy, 2023).

This paper introduces a benchmark derived from 113 FMWC Excel challenges, transformed into a standardized JSON format suitable for programmatic evaluation of LLM performance (Noever & McKee, 2025). The challenges span diverse categories including pattern recognition, financial calculations, game simulations, and data processing tasks. By adapting these challenges from their original spreadsheet format into structured question-answer pairs, we create a benchmark that maintains the essence of the original problems while enabling automated assessment of LLM responses.

Our study evaluates several leading LLMs on this benchmark, analyzing their performance across different challenge categories and identifying specific strengths and weaknesses. The results provide insights into how current models handle tasks that combine numerical reasoning with contextual understanding and rule application—skills that are fundamental to many business applications but underrepresented in existing benchmarks.

The Alpha Excel challenge benchmark represents a step toward more application-oriented AI evaluation, potentially offering a new standard for assessing AI capabilities in business contexts similar to how chess and Go have served as milestones for strategic reasoning (Silver et al., 2017). Unlike these abstract games, however, Excel proficiency has direct real-world applicability, making this benchmark relevant for practical AI applications.

**METHODS**

*Dataset Construction.* The foundational component of our benchmark required an extraction and transformation of challenges from the competitive Excel workbooks. We selected 113 distinct Excel challenges from historical FMWC competitions, ensuring representation across the diversity of problem domains that characterize professional Excel usage (Financial Modeling World Cup, 2024). These challenges originally served as competitive battlegrounds where human Excel experts demonstrated their problem-solving skills within time constraints, typically spanning from 30 minutes to 2 hours per challenge, with their performances often broadcast to audiences in live esports-style events that have gained significant cultural traction (LaurenceLau1, 2022).

The adaptation of these challenges for LLM evaluation demanded a methodology that preserved the core of each problem while transforming its representation. We developed a JSON schema to encapsulate the nature of each challenge, including data, images, and multi-tiered questions. This schema captures the challenge metadata including identifiers, titles, categorical classifications, and difficulty assessments; the problem statement articulating the central task; input data mirroring what human competitors would receive; and structured questions paired with expected solutions that serve as evaluation anchors for LLM responses. Appendix A shows examples of extracted samples, including hard problems that are easy for humans but challenging for LLMs such as letter counting in this benchmark.

This translation process required balancing between preserving problem complexity and eliminating Excel-specific implementation details that would obscure rather than illuminate LLM reasoning capabilities (Noever & McKee, 2025). We devoted attention to maintaining the cognitive challenges while reformulating them in ways that would enable fair and consistent programmatic evaluation. The transformation process involved determining which aspects of the original Excel-specific formulations were incidental to the implementation rather than central to the problem's demands.

The resulting dataset offers an initial distribution of challenge difficulties, from entry-level problems accessible to novice Excel users to advanced challenges that would challenge even decorated FMWC champions (Noever & McKee, 2025; Appendix A). This selection enables analysis of LLM performance across different complexity levels. The composition of the Alpha Excelchallenge set mirrors the eclectic nature of professional spreadsheet usage, comprising approximately 35% financial modeling exercises that test business calculation acumen, 40% game simulation problems requiring complex state tracking and rule application, 15% data analysis challenges focused on pattern discovery and insight generation, and 10% miscellaneous Excel-specific tasks that test technical knowledge of platform capabilities.

*Benchmark Implementation.* The benchmark required development of an evaluation infrastructure capable of presenting Excel challenges to language models in a consistent, interpretable manner while capturing their responses with sufficient fidelity for meaningful comparison. We designed and implemented a standardized and open-sourced evaluation framework that addresses multiple aspects of the assessment process. The framework incorporates a public challenge presentation protocol that transforms the JSON-encoded problem representations into natural language prompts structured to maximize model comprehension while minimizing potential biases. This protocol ensures that each model receives identical problem formulations, establishing a level competitive arena when comparing different LLM models.

The evaluation system employs automated answer processing mechanisms that accommodate the inherent variability in how language models express solutions. Our approach recognizes that semantically equivalent answers may appear in syntactically diverse forms, requiring both exact matching for precision-critical elements and semantic evaluation for conceptual responses. We developed normalization procedures for numerical answers, textual responses, and formula expressions to ensure fair comparison across different model output patterns.

Central to this methodology is a scoring system that avoids binary correctness assessments. This system allocates points according to solution quality, with provisions for partial credit that reflects the degree of correctness in multi-step problems. The scoring rubric differentiates between minor calculation errors and fundamental conceptual misunderstandings, providing a more faithful representation of model capabilities than simplistic correct/incorrect dichotomies.

To ensure scientific reproducibility and facilitate future research extensions, we engineered a comprehensive testing harness that orchestrates the entire evaluation process. This infrastructure automates challenge delivery to model APIs with appropriate rate limiting and error handling, systematically processes response streams, applies the scoring logic, and maintains detailed logs of all interactions. The testing framework maintains consistent environmental conditions across evaluations, including uniform temperature settings, identical few-shot examples, and standardized timeout parameters to ensure fair comparisons between model performances.

*Model Selection and Testing Procedure.* The selection of language models for evaluation required careful consideration of architectural diversity, performance capabilities, and representativeness within the current AI landscape. Our research examined thee contemporary language models that embody different architectural approaches, training methodologies, and performance characteristics: GPT-4o-mini representing OpenAI's most advanced general-purpose model; Mistral exemplifying recent advances from research-focused organizations; and Qwen2.5 demonstrating Alibaba Cloud's progress in open foundation models.

We established protocols for model interaction to ensure fair comparison. All models were accessed through their official API interfaces with identical temperature settings of 0.2—a calibration that balances the deterministic precision necessary for numerical correctness with sufficient creative flexibility for novel problem-solving approaches. Zero temperatures can lack synthesis methods required for real world problem-solving while higher unity temperatures induce wild or creative responses. The prompt engineering process incorporated minimal few-shot examples, carefully designed to illustrate expected response structures without contaminating the evaluation with solution patterns that might artificially advantage certain models or problem categories.

The evaluation incorporated randomization controls to mitigate potential ordering biases. Each model encountered the challenge set in a uniquely randomized sequence, preventing systematic advantages from cross-challenge learning or order-dependent performance variations. For multi-question challenges, we presented the complete question set simultaneously, enabling models to develop holistic understanding of problem context before formulating specific answers—mirroring how human competitors would approach these problems with complete visibility of all requirements.

Practical considerations of computational efficiency and evaluation thoroughness led us to implement a response management systems. We established a calibrated timeout threshold of 60 seconds per challenge, balancing the need for model deliberation against the practical constraints of large-scale evaluation. This parameter was determined through preliminary testing that identified diminishing returns in solution quality beyond this threshold. The response evaluation combined automated exact-match verification for straightforward numerical and categorical answers with expert human assessment for responses requiring semantic interpretation, ensuring that models received credit for correct reasoning expressed through diverse linguistic formulations.

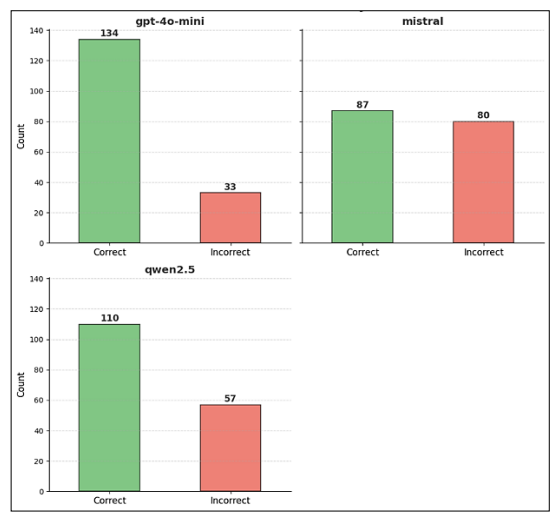

*Figure 1. Performance metrics for correct answers to the Excel benchmarks across open and proprietary*

## RESULTS

*Overall Performance.* The benchmark reveals significant performance variations across different LLMs when tackling Excel challenges. Figure 1 presents the overall accuracy scores for each model, showing the percentage of questions answered correctly across all challenges.

GPT-4o-mini demonstrated the highest overall performance, with particularly strong results on financial modeling and data analysis tasks (>20%). Qwen 2.5 followed closely, showing competitive performance across all categories. Notably, all models performed better on structured financial tasks and data analysis problems compared to game simulation challenges, which often required more complex rule interpretations and multi-step reasoning.

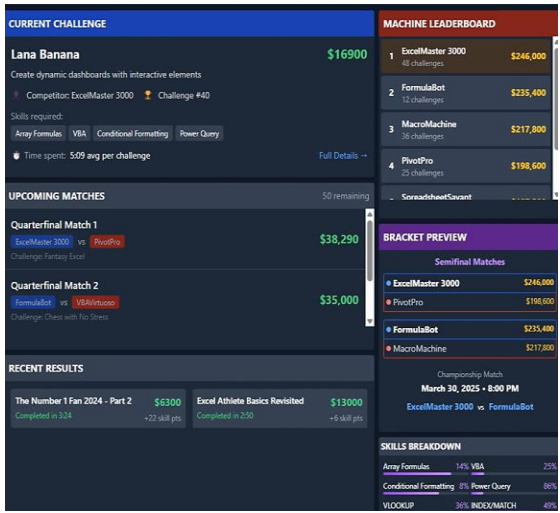

Figure 2. Dashboard tally sheet for competing multiple excel challenges on one large language model

*Performance by Challenge Category.* Analysis of performance across different challenge categories reveals distinct patterns in model capabilities. Figure 2 illustrates the accuracy percentages for each model across major challenge categories as displayed in a real-time dashboard simulation.

The most significant performance variations were observed in game simulation tasks like "Mortal Wombat," "Square of Fortune," and "Ludo," where models needed to track game states and apply complex rule sets. These challenges showed the largest performance gap between the leading models and others, suggesting that sophisticated reasoning about state transitions and conditional logic remains challenging for many LLMs.

Financial modeling tasks showed the most consistent performance across models, possibly due to the more structured nature of these problems and the prevalence of similar examples in model training data. Data analysis challenges, while showing good overall performance, revealed limitations in handling large datasets and complex aggregations.

*Error Analysis.* Understanding the nature of model failures provides crucial insights into limitations and potential avenues for improvement. We conducted error analysis across all model responses, developing a taxonomy of failure modes that emerged from systematic examination of incorrect solutions. This analysis revealed patterns in how language models approach excel-based reasoning tasks and where their cognitive processes diverge from successful problem-solving paths. Figure 3 shows the evaluation of problem difficulty compared to failure rates. These failure categories included calculation mistakes, arithmetic errors, or flawed mathematical operations that produced incorrect values despite sometimes correct solution approaches. Such errors frequently occurred in challenges requiring extended calculation chains where small errors propagated through subsequent steps, culminating in significantly divergent final answers. This finding suggests limitations in how current language models track precise quantitative values through extended reasoning processes.

Rule application errors occurred when models correctly understood problem statements but inconsistently applied or occasionally disregarded specified rules and constraints. These errors were particularly prevalent in game simulation challenges where models needed to simultaneously track and apply multiple conditional rules governing state transitions or scoring mechanisms. The frequency of such errors indicates potential weaknesses in how language models maintain coherent application of complex rule systems throughout extended reasoning chains. The token prediction may not span multi-step applications without a starting solution outline.

Context misunderstanding errors stem from fundamental misinterpretations of problem requirements or constraints. These errors manifested as solutions that addressed different problems than those presented, suggesting limitations in models' abilities to extract precise task specifications from natural language descriptions, particularly when those descriptions contained technical terminology or domain-specific concepts that required contextual interpretation. Formula generation errors represent cases where models correctly conceptualized solution approaches but produced syntactically incorrect or logically flawed formula expressions. These errors highlighted limitations in translating conceptual understanding into precise symbolic representations, particularly for complex nested expressions or those requiring specific Excel function knowledge.

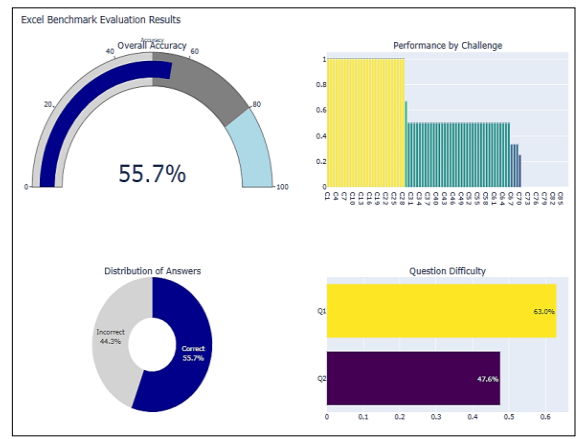

Figure 3. Overall accuracy gauge, performance by challenge, distribution pie chart and question difficulty in the Excel Benchmark

GPT-4o-mini demonstrated notably fewer numerical reasoning errors compared to other evaluated models, suggesting more robust mathematical capabilities within their architectural designs. However, even these leading models exhibited significant difficulties with complex multi-step calculations, particularly those requiring precise tracking of state changes across multiple iterations or those involving conditional logic with numerous branching possibilities (Patel, 2024).

Rule application errors occurred with greatest frequency in game simulation challenges, where successful solutions required models to simultaneously track and apply multiple interdependent rules governing state transitions, scoring mechanisms, and conditional logic. The prevalence of context misunderstanding errors across all challenge categories points toward a universal opportunity for improving how language models extract precise requirements from natural language problem descriptions—a capability that would benefit performance across all problem types and domains.

**DISCUSSION**

*Comparison to Human Performance.* When comparing LLM performance to human benchmarks from FMWC competitions, several interesting patterns emerge. While top human competitors still outperform the best LLMs on complex game simulation challenges by significant margins, the gap is narrowing for structured financial and data analysis tasks (Stradling, 2023). The human advantage is most pronounced in challenges requiring intuitive understanding of problem structure and creative solution approaches. A random selection of Excel challenges might settle on 25% correctness, where a daily Excel user with training might reach 60% proficiency. The world Excel champion scores 85% compared to the best LLM at 80%. Human competitors excel at quickly identifying the core problem patterns and applying targeted solution strategies, while LLMs often attempt more comprehensive approaches that explore multiple potential solution paths (Ngai, 2023). However, LLMs demonstrate advantages in formula recall and consistency. Once a model correctly understands a problem, it rarely makes the kind of simple calculation errors that human competitors might commit under time pressure (Jelen, 2022). This suggests a complementary relationship between human and AI approaches to Excel challenges, where each brings different strengths to the problem-solving process. There remain significant potential for integration AI assistants into "low-code" spreadsheet applications.

*Implications for LLM Development.* The performance patterns observed across our benchmark evaluations illuminate several critical capability gaps in current language model architectures when confronting Excel-type challenges. These findings carry significant implications for future development trajectories that aim to enhance model utility in business contexts.

Numerical reasoning capabilities represent perhaps the most fundamental limitation revealed by our study. Throughout the challenge set, models demonstrated inconsistent precision in mathematical operations, particularly when calculations required maintaining accuracy across multiple interdependent steps. Early LLMs behaved proficiently as "liberal-arts majors" but could often mimic failed math solutions without self-editing. The frequency of numerical errors increased dramatically with calculation complexity, suggesting that current approaches to embedding mathematical reasoning within language models lack the reliability required for mission-critical business applications. Future LLM architectures would benefit from specialized components dedicated to precise calculation verification and validation, potentially incorporating symbolic mathematics engines that can verify computational steps while preserving the flexible reasoning capabilities that language models bring to problem interpretation.

State tracking deficiencies became particularly evident in game simulation challenges where models needed to maintain accurate representations of evolving scenarios through multiple state transitions. While human Excel experts excel at mentally tracking state changes as they modify spreadsheet values, language models showed marked degradation in accuracy as simulation steps accumulated. This finding suggests an architectural limitation in how current models maintain and update contextual information through extended reasoning chains. In other contexts, this deficiency includes the LLM's failure to plan. Future developments might explore dedicated memory mechanisms or specialized attention architectures that prioritize state consistency throughout multi-step processes, potentially drawing inspiration from how human working memory operates when manipulating symbolic representations.

Rule synthesis presented another significant challenge, with models failing to integrate multiple conditional rules into consistent solution strategies. When faced with complex rule systems governing Excel challenge scenarios, models often applied individual rules correctly in isolation but struggled to maintain logical consistency across rule interactions. This finding suggests opportunities for enhancing how models represent and reason about rule systems, potentially through architectural components specifically designed to maintain logical consistency or through training approaches that emphasize detecting and resolving rule conflicts in complex scenarios. This deficiency equates to both LLM failure to plan but also to show real-world commonsense or reflect expected physics.

Domain-specific knowledge representation varied considerably across models and challenge categories. While some financial modeling tasks benefited from domain knowledge embedded in model training data, this knowledge appeared inconsistently distributed, creating performance disparities across related problems. This observation suggests value in specialized fine-tuning approaches that systematically incorporate domain expertise into model capabilities, particularly for high-value business domains where Excel serves as a primary analytical tool. Examples of a failure to interpret multi-step unit conversions feature in FMWC challenges, including rank orders, mixed English or metric units and currency conversion.

In total, these benchmark results provide direction for future language model development efforts, particularly for models intended to assist with business analysis and financial modeling tasks. The Alpha Excel benchmark establishes a practical evaluation framework for measuring progress in these capabilities, offering a bridge between abstract reasoning benchmarks and real-world business applications (Microsoft Research, 2022). Like chess or Go, the inevitable comparisons between human champions and machine models assist in tracking LLM progress. By addressing these specific limitations, future models could dramatically enhance their utility as collaborative partners in business analysis contexts, potentially transforming how professionals interact with quantitative information and decision support systems.

*Benchmark Limitations.* Despite its uses in assessing language model capabilities on business-relevant tasks, this Excel benchmark methodology suffers several inherent limitations that merit acknowledgment and consideration in interpreting results. The translation from interactive spreadsheet challenges to text-based evaluations necessarily sacrifices elements of the original task environment. This conversion process, while enabling programmatic evaluation, diminishes the rich interactive dimensions of spreadsheet work where visual scanning, cell navigation, and formula manipulation constitute integral parts of problem-solving. The resulting abstraction potentially underrepresents certain cognitive demands present in authentic Excel usage, particularly those related to spatial reasoning and interactive debugging processes. Python libraries like docling (IBM) assist in more literal renderings from PDF and Microsoft Office products.

Our evaluation metrics prioritize solution correctness as the main performance indicator, a necessary simplification that potentially obscures valuable insights about problem-solving approaches. This outcome-focused assessment does not capture the efficiency, elegance, or explainability of solution strategies—qualities highly valued in professional contexts. Human Excel experts often distinguish themselves not merely through correct answers but through solution paths that demonstrate conceptual clarity and maintainability. The benchmark's emphasis on final outputs rather than solution trajectories limits our ability to evaluate these qualitative dimensions of performance. Current LLM methods that document "chain of thought" or "deep research" exist to provide explanations and partial credit based on reasoning steps.

Temporal dynamics represent another area where our benchmark diverges from authentic Excel competitions. The original FMWC challenges incorporate calibrated time constraints that create strategic pressure on human competitors, forcing trade-offs between solution completeness and deadline adherence. Our implementation relaxes these constraints to accommodate the fundamentally different processing characteristics of language models, establishing a uniform 60-second response window that bears little resemblance to the strategic time pressures facing human competitors. This methodological adaptation, while necessary for practical evaluation, removes an important dimension of authentic Excel competition.

The standardized JSON representation format employed throughout our benchmark sacrifices important visual and spatial aspects of spreadsheet work. Excel experts typically leverage spatial organization, cell formatting, and visual cues to structure problems and solutions—contextual dimensions entirely absent from our text-based challenge representations. This limitation is particularly significant for problems where spatial relationships between data elements provide important solution cues or where visual pattern recognition constitutes a primary cognitive pathway to insight.

These methodological limitations suggest practical directions for future benchmark refinements. The incorporation of more interactive evaluation protocols could better capture the dynamic nature of authentic Excel problem-solving. Process-oriented assessment metrics might provide more nuanced insights into solution quality beyond binary correctness evaluation. The development of evaluation frameworks that incorporate visual and spatial dimensions of spreadsheet work would create more authentic assessment contexts. Such enhancements would move Excel benchmarking closer to the validity needed to fully understand how language models might complement human expertise in real-world business analysis tasks.

**FUTURE WORK**

The research open several promising future directions from this benchmark study. First, incorporating challenges from recent FMWC competitions would create a continuously evolving evaluation framework that tracks language model improvements against the advancing frontier of human Excel championships. LLMs could earn a seat at the table in FMWC challenges. This dynamic approach would prevent benchmark saturation while maintaining relevance to contemporary business analytics practices.

Specialized challenge tracks focusing on domains of Excel work represent another valuable extension. Targeted evaluation suites for financial modeling could assess capabilities relevant to investment analysis and corporate finance, while specialized tracks for statistical analysis, data transformation, and visualization would illuminate model performance in other critical business applications. These domain-specific extensions would provide more granular insights into how model capabilities align with specialized professional needs. For instance, the sometimes maligned Visual Basic Macro language of Excel shows potential for LLM automation in ways that other foundational models (>30 billion parameters) can master programming languages like JavaScript, Python, and Web 2.0 frameworks.

The development of adaptive challenge sets with progressive difficulty scaling would enhance benchmark sensitivity, particularly for identifying precise capability thresholds in high-performing models like GPT-4o-min. Such adaptive evaluation approaches could incorporate automated difficulty calibration based on model performance patterns, creating more efficient assessment protocols that quickly identify specific capability boundaries. This approach would be particularly valuable as model performance improves, preventing ceiling effects from obscuring meaningful performance differences.

Secondly, the transition toward more interactive evaluation methods would better capture the collaborative nature of real-world Excel usage. One notable attractive feature of spreadsheets is their shareability for team contributions. Development of simulated Excel environments where models can directly manipulate data representations would create a more authentic assessment context, measuring not just solution quality but also the efficiency and elegance of solution paths—qualities highly valued in professional Excel work.

Multi-turn evaluation protocols represent a particularly promising direction, enabling models to receive feedback and incrementally refine solutions. Like current agentic AI, such approaches would better mirror how human experts develop Excel solutions through iterative refinement, potentially revealing different capabilities than those observed in single-turn evaluations. This interaction could also illuminate how well models incorporate feedback to improve solution quality, a critical capability for collaborative business applications.

Collaborative benchmarks assessing how effectively models assist human users with Excel tasks would provide perhaps the most ecologically valid performance metrics. Such evaluations would measure not just model performance in isolation but the quality of human-AI collaboration outcomes—ultimately the most relevant metric for practical applications. These studies could incorporate diverse user expertise levels to understand how model assistance benefits different user populations, from Excel novices to seasoned professionals.

Thirdly, development of specialized fine-tuning datasets focused on Excel problem-solving could address the error patterns observed in our initial benchmark. Such datasets could emphasize numerical precision, state tracking, and rule application—providing targeted training signals for capabilities that showed consistent weaknesses across models.

Architectural innovations represent another promising direction, particularly hybrid approaches that combine language model capabilities with specialized numerical reasoning components. Such hybrid architectures could leverage the strengths of language models in context understanding and problem interpretation while addressing their weaknesses in precise calculation through dedicated numerical processing modules. This approach might draw inspiration from how human experts combine conceptual and procedural knowledge when solving spreadsheet problems.

Curriculum learning strategies could enhance training efficiency by progressively introducing spreadsheet challenges with increasing complexity. Such approaches would first establish foundational capabilities like basic formula understanding before advancing to complex multi-step calculations and rule systems. This progressive skill development might more effectively build the capability hierarchies needed for advanced Excel problem-solving, potentially requiring less training data than approaches that immediately target complex problems.

Fourthly, this benchmark has natural extensions to other software domains that feature complex problem-solving. Adapting our approach to database query challenges would create valuable evaluation frameworks for SQL capabilities, an adjacent skill domain with similar relevance to business analytics. Such adaptations would require careful consideration of how to maintain problem authenticity while enabling programmatic evaluation. Similar benchmarking approaches could illuminate model capabilities with data visualization tools, another critical component of the business analytics ecosystem. Challenges in this domain would assess how effectively models can translate data insights into appropriate visual representations, testing both analytical and communication capabilities simultaneously.

The extension of our evaluation methodology to other business software applications would provide a more comprehensive assessment of how language models might transform professional workflows across the enterprise software landscape. Such cross-domain benchmarks could reveal whether capabilities transfer effectively across application contexts or whether domain-specific limitations require targeted development efforts. These diverse research directions collectively aim to bridge the persistent gap between abstract AI benchmarks and practical business applications, providing both academic researchers and commercial developers with insights needed to enhance language model utility in real-world contexts (Pluss et al., 2022; Nagdeote et al., 2023). By addressing the limitations revealed in our current benchmark, future research can accelerate the development of AI systems that meaningfully complement human expertise in business analytics.

**CONCLUSIONS**

This paper introduces a novel benchmark for evaluating LLM capabilities using challenges derived from the Financial Modeling World Cup Excel competitions. By converting 113 diverse Excel challenges into a standardized JSON format, we create a practical and open-source assessment framework that tests multiple dimensions of model performance on business-relevant tasks.

Our evaluation of three leading LLMs reveals significant performance variations across models and challenge categories, with financial tasks generally showing higher accuracy than game simulations. The best-performing models demonstrate promising capabilities on structured problems but still lag behind human experts on complex challenges requiring intuitive understanding and creative problem-solving (Horiuchi, 2017; Microsoft, 2017).

The Excel benchmark represents a step toward more application-oriented AI evaluation, focusing on skills with direct business relevance rather than abstract reasoning alone. This approach helps bridge the gap between academic AI research and practical business applications, potentially accelerating the development of AI systems that can meaningfully assist with real-world analytical tasks.

As LLMs continue to evolve, benchmarks like this one will play an important role in guiding development toward capabilities that combine the strengths of machine learning with the practical needs of business users. The goal is not to displace human Excel expertise but to augment it with AI capabilities that handle routine aspects of analysis while enabling humans to focus on interpretation and decision-making (Esra, 2024; Jenny et al., 2021).

## ACKNOWLEDGEMENTS

We would like to thank the Financial Modeling World Cup organization for providing the rich repository of Excel challenges that made this research possible. Special thanks to Andrew Grigolyunovich and the FMWC team for their pioneering work in establishing Excel as a competitive discipline. We also acknowledge the support of the PeopleTec Technical Fellows program for the research support that enabled the extensive model evaluations conducted in this study.

## Appendix A: Example Excel FMWC Interpreted Skills

**Tough LLM Letter Counting**

```
"challenge_id": "09-Letter-Soup",
        "title": "Letter Soup
Challenge",
        "category": "Text Processing",
        "difficulty": "Intermediate",
        "problem_statement": "Given a
string of letters, count the
occurrences of a specified character.
Some questions may involve case
sensitivity or transformations.",
        "data": {
            "letter_sequences": [

"uzTyjLmnfIqULrWCoArlPvoLzKNVCzTpIMoltr
EnLcDpS"
            ],
            "target_characters": [
                "L"
            ]
        },
        "questions": [
            {
                "question_id": "Q1",
                "text": "How many times
does the displayed 'Character' appear
in the given text?",
                "input_data": {
                    "text":
"uzTyjLmnfIqULrWCoArlPvoLzKNVCzTpIMoltr
EnLcDpS",
                    "character": "L",
                    "case_sensitive":
true
                },
                "expected_answer": 4
            },
```

**Optimization Question and Answer**

```
        "challenge_id": "Share-the-
Ride",
        "title": "Carpool
Optimization",
        "category": "Logistics
Optimization",
        "difficulty": "Advanced",
        "problem_statement": "Given a
list of ride requests, destinations,
and driver availability, optimize the
route to minimize total travel
distance.",
        "data": {
            "ride_requests": [
                {
                    "passenger":
"Alice",
                    "pickup": "A",
                    "dropoff": "C"
                },
                {
                    "passenger": "Bob",
                    "pickup": "B",
                    "dropoff": "D"
                }
            ],
            "driver_availability": [
                "Driver 1",
                "Driver 2"
            ]
        },
        "questions": [
            {
                "question_id": "Q1",
                "text": "Which driver
should pick up Alice to minimize travel
distance?",
                "expected_answer":
"Driver 1"
            },
            {
                "question_id": "Q2",
                "text": "What is the
total optimized route distance?",
                "expected_answer": 15.4
            }
        ]
```